# Introducing Construct Theory as a Standard Methodology for Inclusive AI Models

**Authors: Susanna Raj, Sudha Jamthe, Yashaswini Viswanath, Suresh Lokiah**

**Abstract**

Construct theory in social psychology, developed by George Kelly are mental constructs to predict and anticipate events. Constructs are how humans interpret, curate, predict and validate data; information. AI today is biased because it is trained with a narrow construct as defined by the training data labels. Machine Learning algorithms for facial recognition discriminate against darker skin colors and in Joy Buolamwini's ground breaking research papers, the inclusion of phenotypic labeling is proposed as a viable solution. In Construct theory, phenotype is just one of the many sub-elements that make up the construct of a face. In this paper, we present 15 main elements of the construct of face, with 50 sub-elements and tested Google Cloud Vision API and Microsoft Cognitive Services API using FairFace dataset that currently has data for 7 races, genders and ages, and we retested against FairFace Plus dataset curated by us. Our results show exactly where they have gaps for inclusivity. Based on our experiment results, we propose that validated, inclusive constructs become industry standards for AI/ML models going forward.

## Introduction

AI is biased because it is built on biased data, but AI ethics experts on the other hand exhort that we human beings are biased (Kroenung et al. 2022), therefore human data is biased, so bias cannot be mitigated or eliminated unless historical and social biases in society are eliminated (Vlasceanu et al. 2022). Since this is not possible within a specified timeline, the next best thing is to have representation of all groups at every stage (Cogwill et al. 2020) and in every dataset to mitigate biases (Tommasi et al. 2015).

This is the cornerstone of Inclusive AI, but a critical perspective missing from this type of bias mitigation is that equal representation alone of all groups will not solve the problem unless each group understands how the "other" is perceived from within and perceives them and others from outside; historically, culturally, and cognitively. For example, given that lighter skin color is held in high preference explicitly and implicitly in society, that even those with darker skin colors show this preference and social acceptance for lighter skin tones (Maxwell et al. 2014). Societal biases define personal biases (John Murray, 2019), and they are all built from the mental constructs that we hold for everything including skin color.

Inclusive AI, free of bias, cannot be built unless we understand that the data we create as humans and collect from the day we are born, holds different meanings; constructs for all of us. Constructs are an integral part of natural, cognitive, and social sciences. For example, the natural sciences use constructs like gravity and temperature to help us understand natural phenomena that we can feel, perceive but not see. In cognitive science, we use constructs for abstract phenomena like intelligence, extroversion, introversion, control, confidence, power and many more intangible yet valid behavioral entities. Construct theory says all attributes of the construct form the construct in our mind. All attributes that do not fit in with the construct also define the construct. The boundaries of what does not fit into the construct define its edges.

In social psychology, construct theory was developed by George Kelly (George Kelly, 1991) which says that we experience the world through different lenses of our constructs. These constructs are used to predict and anticipate events, which in turn determines our behaviors, feelings, and thoughts in interpreting them. We test and retest our constructs until it proves its validity across events. The constructs of one generation, culture, or language varies but when constructs are thoroughly defined, clearly articulated to the point that different groups of people can agree that it means the same or similar thing, it becomes a commonly accepted instrument and metric to understand and communicate the phenomena.

In simpler terms, constructs are what humans use as prediction models.

Anticipation and prediction of our responses to stimuli from others, responses of others to different stimuli from us, schematic prediction of environments and events are all based on our mental personal constructs. This is how human intelligence creates and curates the data around us. Artificial Intelligence on the other hand is a broad discipline of statistical algorithmic applications, models that are built with human data: that are intelligent enough to predict, enhance and support human tasks. AI is built by operationalizing variables at the conceptual stage to define its scope and impact. Without understanding how humans perceive both tangible and intangible concepts, variables are defined by the AI developers who will, unbeknownst to themselves, be imposing their own construct definitions into the model. (R. Mohanani et al. 2020)

AI is still built with construct definitions but of one group of humans only, from a selective community. Including other groups or simply adding more training data to the model will not solve this problem when the variable measured is not defined inclusively. For example, the construct of alertness was defined as eyes open which led to Asian drivers being wrongly identified as sleeping or drowsy (Cheng, Selina, 2016). Researchers recently showed that by defining it with more inclusive construct definitions (Anonymous et al. 2022), this problem could be solved even without adding additional training data.

A valid construct definition before we build an operational definition makes a significant contribution in research, i.e., the accuracy of the predictions depends largely on the quality of the conceptual, observational and practical foundations of the construct in question – whether the operational definition of the variable meets or holds water to its theoretical meaning.

There is a huge lack of understanding about its importance in AI models. Research has shown that using inclusive constructs as a labeling approach increased machine learning model performance. (Anonymous et al, 2021) As the industry faces tremendous scrutiny for the various biases in AI applications, it is extremely crucial to go back to basics and look at this problem from the lens of cognitive and social science methodologies.

Construct definitions take the complexity of anything, directly observable or not and provide a common but measurable meaning to all variables involved, end to end. (Anonymous, 2021)

This paper proposes the use of natural cognitive processing, NCP which relies entirely on personal and mental construct theory to make sense of the world around us and in the production of human data, as a framework model to make AI more inclusive and less biased. We do this by building out a construct definition for an observable variable; face and testing top facial recognition algorithms against it.

## Use of construct theory to operationalize the variable "face" as a construct

Facial recognition models have come under public scrutiny for discriminating against Black, Asian races and other minority populations. A study of facial recognition models by NIST (Patrick Grother et al. 2020) found that 189 algorithms showed racial bias against women of color. The prevailing solution for this, in the minds of most data scientists and AI/ML engineers is to add more of the missing data. If women of color are missing in the training data, leading to a biased model outcome, then more training datasets of them are added as direct bias mitigation offset. This method does not make the model more inclusive or less biased, it simply mitigates one of the highlighted bias while still holding on to its discriminatory impact which included a non-exhaustive, non-inclusive construct definition of the key variable; face.

In this paper, we present 15 main elements of the construct of face, with 50 sub-elements (this is clearly not the most complete or even exhaustive definition but given the limitation of time, we have limited the scope for this paper). Construct theory proposed by George Kelly was built on one Fundamental Postulate and 11 Corollaries. 1.The Construction Corollary. 2. The Experience Corollary. 3. The Individuality Corollary. 4. The Organizational Corollary. 5. The Dichotomy Corollary. 6. The Choice Corollary 7. The Range Corollary 8. The Modulation Corollary. 9. The Fragmentation Corollary. 10. The Commonality Corollary 11. The Sociality Corollary.

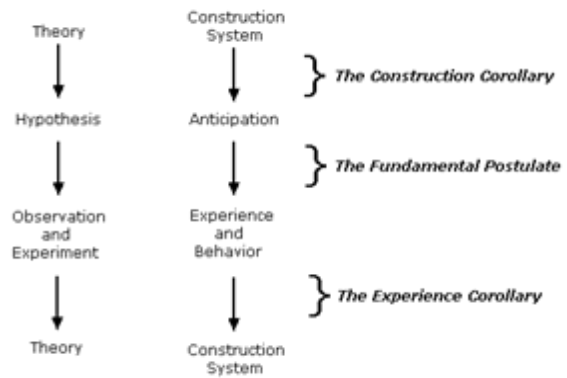

Figure 1. shows the postulate and two corollary connections in this theory.

(Kelland, Mark D, 2022) **The Fundamental Postulate**: A person's processes are psychologically channelized by the ways in which he/she anticipates events. **The Construction Corollary**: We conservatively construct anticipation based on past experiences. **The Experience Corollary**: We learn new information, adapt, relearn, and reconstruct to improve prediction and control. **The Individuality Corollary**: Persons differ from each other in their construction of events. **The Organizational Corollary**: We develop our constructs in a systemic and organized way with some constructs ranked higher than others. **The Dichotomy Corollary**: We construct systems composed of a finite number of dichotomous constructs, of both positive and negative aspects. **The Choice Corollary**: We choose the best construct within the dichotomy that offers the greater possibility for extension and definition of our system. **The Range Corollary**: Constructs have a finite range of application within the system. **The Modulation Corollary**: Some our constructs are more easier to modify than others. **The Fragmentation Corollary**: We may employ or tether to constructs that do not fit well in our predictions to adjust for temporary conflicts. **The Commonality Corollary**: Some constructs are shared by those with similar lived experiences and do have a commonality in psychological processes. **The Sociality Corollary**: Constructs of one person or society help define the constructs of another person or society.

**Experiments Supporting data**

Our hypothesis is that we need the construct framework to make AI inclusive and the approach of training data with a diverse dataset by including diversity for race, gender and age does not work in building Inclusive AI. The construct framework is not about addition of missing elements as is the general practice of bias mitigation today. The construct framework is about inclusion of all possible attributes that form the variable instead of just adding a singular attribute. As mentioned in our paper, Construct theory follows 11 corollaries. Our experimental design was based on the *Organizational Corollary*, in which constructs are connected to one another in hierarchies and network of relationships.

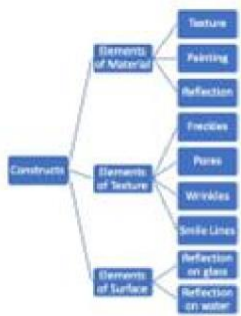

Figure 2: Constructs shown in their hierarchical arrangement of elements.

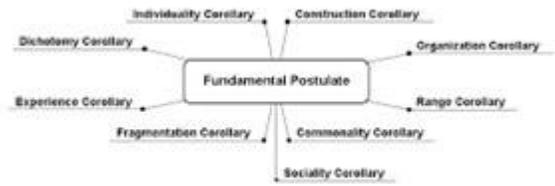

Figure 3: Construction Corollary with Fundamental Postulates and 11 Corollaries by George Kelly:

What can construct definition do to help biases seen in facial recognition models? Taken in this construct framework method, facial recognition models before they even start the data collection would have had a library of construct definitions for a face. That construct definition would have had categories for shapes, sizes, shades, light, age, context, culture and place in which a face appears. Under each category, for example skin tone alone, there has been research done by the non-tech sector that has identified the possibilities of up to 1,866 colors and 4,000 possible shades. (Cascone, Sarah, 2018)

Under shapes, not only shapes of the overall face would have been categorized, but also the overall shapes of eyes, eyelids, noses, ears and mouths belonging to all races would have been tagged.

Example of a well-defined construct of a face using organizational corollary of Construct Theory.

- Angle - vertical
- Angle - Horizontal
- Angle - upside down
- Angle - 45 degree right
- Angle - 45 degree left
- Shape - Oval
- Shape - Round
- Shape - Square
- Shape - Diamond
- Shape - Heart
- Shape - Pear
- Shape - Oblong
- Sizes - Narrow
- Size - Wide
- Shade - Shadows
- Skin tone - total of 10
- Lighting - Bright

- Light - Blur
- Light - Poor lighting
- Race - White
- Race - Asian

- Race - Black
- Race - Indigenous
- Age - baby
- Age - Teen
- Age - Youth
- Age - Adult
- Age - Senior
- Context - Background
- Culture - Asian
- Culture - African
- Culture - Piercing
- Culture - Head Covering
- Material - texture
- Material - painting
- Material - reflection
- Texture - freckles,
- Texture - pores
- Texture - wrinkles
- Texture - smile lines
- Surface - reflection on glass
- Surface - reflection on water
- Disabilities - burns
- Disabilities - tattoos
- Disabilities - painting
- Face Covering - Oxygen Mask
- Face Covering - Glasses
- Face Covering - Beard
- Face with makeup
- Parts of a face

**Experiment Results**

We tested Google and Microsoft against synthetic data for elements of constructs for faces and found the following results showing that both algorithms failed for several elements of constructs. This shows lack of inclusivity for several people and situations which can be supported by included construct elements in the training of the algorithms.

For Race, FairFace has a Black category. But when we tested with Race- African faces, Google did not return any data and Microsoft reduced the face detection confidence down to 82.18%. Google also has low confidence for faces with face painting. Both algorithms struggle with African cultural tattoos on faces. They both do well with several types of Black hairs such as crown and curls but Google face detection fails at 47% for hair with braids.

## Our Experimental Approach and results

**Experiment 1**
We took the FairFace dataset which is the most diverse face open dataset which has diversity included for 7 races, genders and ages and tested it for fairness for 50 elements and sub elements of our 15 constructs of face and the results showed it failed for 50% of all the categories.

Since such a fair and diverse dataset is what top face detection algorithms are trained on, we tested Google Cloud Vision API and Microsoft Cognitive Services API using FairFace dataset. FairFace in their paper report results for top four Face detection API (Amazon, Microsoft, IBM & Google) show 74% to 99% performance on detecting faces inclusively.

**Experiment 2**
We ran tests of Google Cloud Vision API and Microsoft Face API against the FairFace+ dataset which brings a richer set of constructs to go beyond race, gender and age to add more dimensions of humans to identify human faces.

**Experiment Results/Insights**
Google failed with a zero for African ethnicity faces though it performed well for Black category of faces from FairFace. Microsoft on the other hand performed well for African ethnicity faces but failed when these faces have face covering or marks or disability or cultural marks. Both algorithms performed well for White faces but when tested for White faces with a rich construct set they failed detecting the same white faces.

**Abnormal results due to hidden biases**

Microsoft's Cognitive Services API mistook the close-up shot of a human face for a cat.

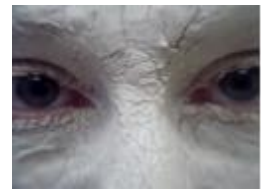

The face had white textured paint on it. Facial paints are part of many cultures and ethnicities. Algorithms trained on just a superficial definition of races hold an implicit bias towards non-western centric cultural identities. The API also misidentified the face of a man covered in soap foam as that of a white and brown rock. Both were a clear failure on one of the elements of our constructs in FairFace + for facial texture.

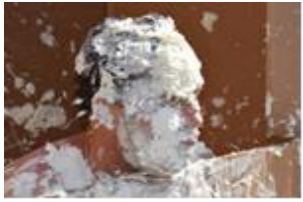

And faces with beards were identified as a person wearing flowers in their hair.

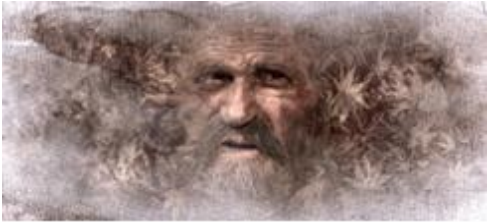

## FairFace+ Open Dataset

We created an open dataset called FairFace+ that includes 30 images that covers all the constructs that failed the FairFace dataset. We added ethnicity, cultural difference of face shapes, sizes, face covering and inclusion of distinctive faces of people with disabilities and burns.

## Sampling the FairFace dataset

We randomly sampled 383 images from the 86k FairFace training dataset (Karkkainen, K., & Joo, J. (2021) . The sample size was chosen based on 95% confidence and 5% margin of error. The sample had images organized by Age, Gender, Race. We wrote a simple fairness analyzer using Python to run this dataset against Google and Microsoft Face APIs and captured the results

## Explainable AI to test our rich construct framework on Google and Microsoft APIs.

We hosted the FairFace+ dataset of 30 images on GitHub and ran Google and Microsoft Face API against each of these images and captured the face detection confidence.

|  | Google Cloud Vision API | | | |
|---|---|---|---|---|
| Race | Black | White | Latinx | Asian |
| Age | Pass | Pass | Pass | Pass |
| Gender | Pass | Pass | Pass | Pass |

Table 1a: FairFace dataset Fairness Analysis with Google Cloud Vision API

|  | Microsoft Azure Cognitive Services | | | |
|---|---|---|---|---|
| Race | Black | White | Latinx | Asian |
| Age | Pass | Pass | Pass | Pass |
| Gender | Pass | Pass | Pass | Pass |

Table 1b: FairFace dataset Fairness Analysis with Microsoft Face API

|  | Google Cloud Vision API | | | | |
|---|---|---|---|---|---|
| Race | Black +African | White | Indigenous | Latinx | Asian |
| Age | F | F | F | F | F |
| Skintone | F | Pass | F | F | F |
| Face shape | F | F | F | F | |
| Shadows | F | F | F | F | |
| Face additions | F | F | F | F | |
| Hair types | F | Pass | F | Pass | |

Table 2a: Construct Mapping with FairFace + dataset for Google API

|  | Microsoft Azure Cognitive Services | | | | |
|---|---|---|---|---|---|
| Race | Black +African | White | Indigenous | Latinx | Asian |
| Age |  |  |  | F |  |
| Skintone |  | Pass |  | F |  |

| Face shape |  |  |  | F |  |
| --- | --- | --- | --- | --- | --- |
| Shadows |  |  |  | F |  |
| Face additions |  |  |  | F |  |
| Hair types |  |  |  | F |  |

Table 2b: Construct Mapping with FairFace + dataset for Microsoft API

## Conclusion:

Construct theory from psychology can help add inclusivity to train AI better. Our preliminary paper results show insights from top face detection algorithms on exactly where they have gaps for inclusivity and can act as an explainable AI tool to improve inclusiveness in face detection algorithms.

Code snippets and media submitted in supplementary.